\documentclass[%
reprint,
twocolumn,
superscriptaddress,
graphicx,
floatfix,
amsmath,
amssymb,
bibnotes,
aps,
prb,
]{revtex4-2}
\usepackage[]{graphicx}
\usepackage{epstopdf}
\usepackage{MnSymbol}
\usepackage{latexsym}
\usepackage{color}
\usepackage{dcolumn}
\usepackage{bm}
\usepackage[mathlines]{lineno}
\usepackage{usebib}

\begin{document}

\preprint{Submitted to Optics Letters}

\title{Designing thermal radiation metamaterials via hybrid adversarial autoencoder and Bayesian optimization}

\author{Dezhao Zhu}
\affiliation{China-UK Low Carbon College, Shanghai Jiao Tong University, Shanghai 201306, China}

\author{Jiang Guo}
\affiliation{Department of Mechanical Engineering, The University of Tokyo, 7-3-1 Hongo, Bunkyo, Tokyo 113-8656, Japan}

\author{Gang Yu}
\affiliation{China Building Materials Academy, State Key Laboratory for Green Building Materials, Beijing 100024, China}

\author{C. Y. Zhao}
\affiliation{China-UK Low Carbon College, Shanghai Jiao Tong University, Shanghai 201306, China}

\author{Hong Wang}
\affiliation{Materials Genome Initiative Center, School of Material Science and Engineering, Shanghai Jiao Tong University, Shanghai 200241, China}

\author{Shenghong Ju}
 \email{shenghong.ju@sjtu.edu.cn}
\affiliation{China-UK Low Carbon College, Shanghai Jiao Tong University, Shanghai 201306, China}
\affiliation{Materials Genome Initiative Center, School of Material Science and Engineering, Shanghai Jiao Tong University, Shanghai 200241, China}

\begin{abstract}
Designing thermal radiation metamaterials is challenging especially for problems with high degrees of freedom and complex objective. In this letter, we have developed a hybrid materials informatics approach which combines the adversarial autoencoder and Bayesian optimization to design narrowband thermal emitters at different target wavelengths. With only several hundreds of training data sets, new structures with optimal properties can be quickly figured out in a compressed 2-dimensional latent space. This enables the optimal design by calculating far less than 0.001\% of the total candidate structures, which greatly decreases the design period and cost. The proposed design framework can be easily extended to other thermal radiation metamaterials design with higher dimensional features.
\end{abstract}

\date{\today}

\maketitle
Thermal radiation metamaterials with excellent wavelength selective spectrum can be realized via reasonable design and have attracted considerable interests in applications of thermophotovoltaics \cite{zhou2016integrated}, incandescent light sources \cite{ilic2016tailoring} and biosensing \cite{luo2016perfect}. With the advances in nano-fabrication technology, metamaterials such as 1D thin film layers \cite{zhao2014ultra}, 2D-grating \cite{dahan2008extraordinary}, and 3D photonic crystals \cite{lin2000enhancement} have been proposed to implement wavelength selective spectrum. However, with the increased degrees of freedom for metamaterial design, the total number of candidate structures will grow exponentially to an enormous space. The intuition-based or trial-and-error methods are undesirable for optimization with high-dimensional and complex objectives.

To overcome the above bottle-neck problem, materials informatics (MI) \cite{ju2019materials, ju2020designing, 10.1088/978-0-7503-1738-2ch5}, which combines data-oriented algorithms and material property characterization either by simulation or experiment, has been proposed to accelerate the material design process. During the past few years, various gradient or meta-heuristic based algorithms have been proposed for non-intuitive metamaterials design, including topology optimization \cite{sell2017large}, adaptive genetic algorithm \cite{jafar2018adaptive} and lazy ants \cite{zhu2019optimal}. When the design degrees of freedom increase to high-dimensional space, these optimization algorithms will suffer from the falling into local optima and depending on large number of iterative computations or experiments. The Bayesian optimization (BO) has been proposed to improve the optimization efficiency and has been successfully applied to the design of narrowband emitter \cite{sakurai2019ultranarrow} and radiation cooling materials \cite{guo2020design}. However, for the cases with large candidate space, it is necessary to divide into subgroups to overcome the BO's shortage of requiring huge computational memory. Compared to BO, quantum annealing (QA) can overcome computational barrier and has been used in the design of radiative cooling metamaterials \cite{kitai2020designing}. Nevertheless, only structures that can be encoded as binary variables are available for optimization, and the maximum design degrees of freedom are dependent on the quantum annealer. Multi-layer perceptron (MLP), as a common deep neural network (DNN), can also achieve optimal design of continuous variables by learning the relationship between inputs and outputs \cite{garcia2021deep}. However, when it is applied to the high-degree freedom design problem, there will be challenges of huge required raining data set, underfitting due to many-to-one problems \cite{liu2018training}, and local optimal solution. Recently, the adversarial autoencoder (AAE) network has been coupled with adjoint topology optimization (TO) technique for optimizing thermal emitters \cite{kudyshev2020machine} by introducing the compact design space representations, which further improves the efficiency of DNN for optimization problems. To better perform the multi-parametric global optimization (GO), the conditional adversarial autoencoder (c-AAE) network coupled with differential evolution (DE) optimizer \cite{kudyshev2021machine} was proposed to optimize the surface topology. However, there is still no efficient design method for multilayer structures with high degrees of freedom, and the search efficiency in the compact design space obtained from trained AAE models can be further improved.

In this letter, we combined the AAE network with BO to design narrowband thermal radiation metamaterials, which balances the exploration and exploitation for fast global search in the compressed latent space with normal distribution. To showcase the performance of the proposed AAE-assisted BO model, we optimized the material selection in multilayer structure that can be represented by binary pink and white pictures for narrowband emission at three different wavelengths. The optimization framework ensures almost ideal thermal emission through calculating the figure of merit (FOM) of each recommended structure, but only far less than 0.001\% of the total number of candidate structures need to be calculated. Design problems with higher degrees of freedom in optics, thermal conductance and thermoelectric are the potential applications of this study.

\begin{figure}[ht!]
\centering
\includegraphics[width=\linewidth]{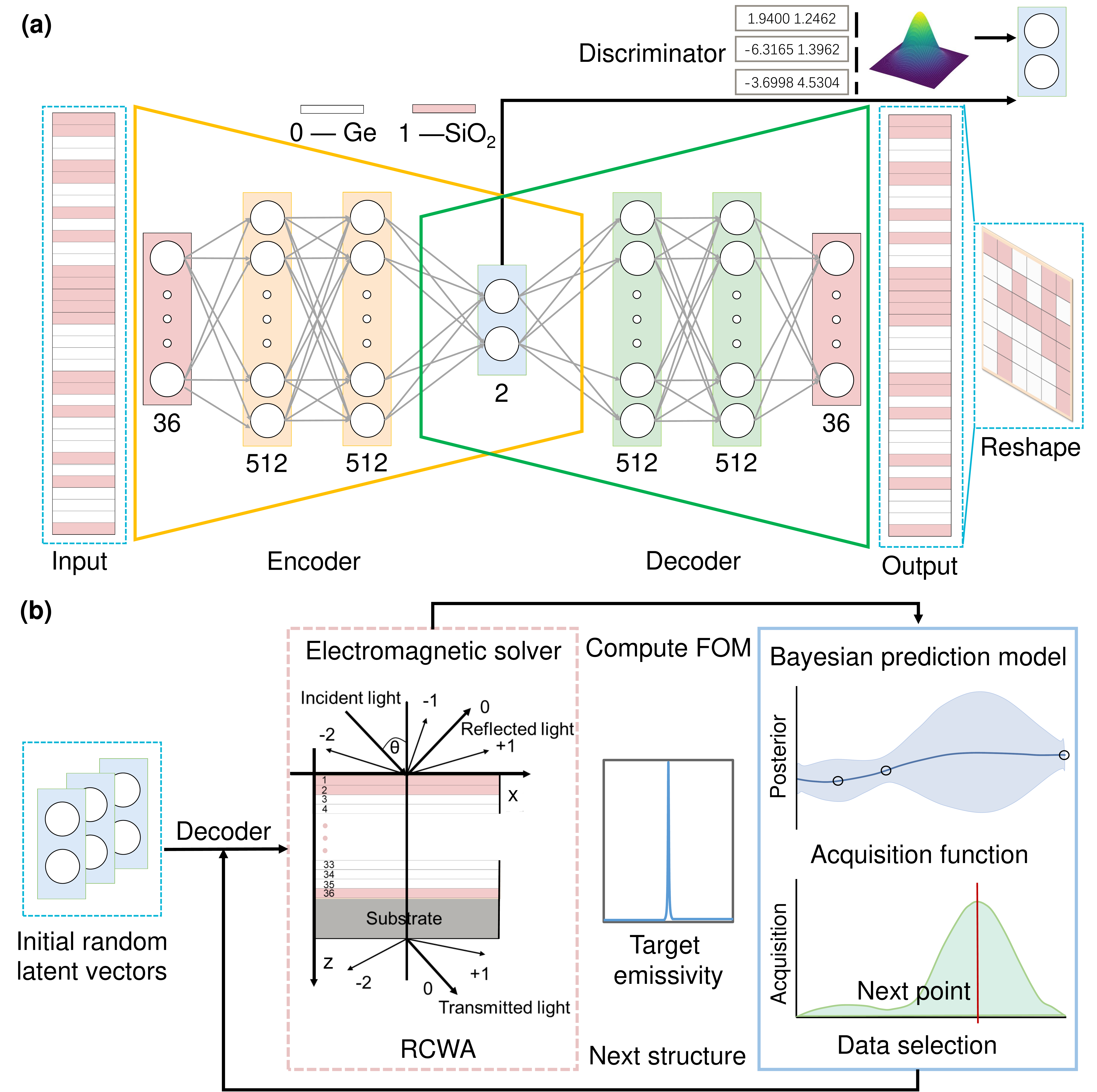}
\caption{Schematics of the hybrid framework for thermal radiation metamaterials design. (a) Schematic of training the AAE model. (b) Schematic of the decoder combining electromagnetic solvers RCWA and BO for global optimization.}
\label{fig:1}
\end{figure}

The schematic of the hybrid optimization framework is shown in Fig. \ref{fig:1}, which combines AAE-assisted BO with the electromagnetic solver rigorous coupled wave analysis (RCWA) \cite{liu2012s4}. The narrowband thermal emitters are composed of 36 cell layers of Ge and SiO$_2$ materials. The materials are chosen from commonly used semiconductors with high refractive index and dielectric materials with low refractive index, while the substrate material uses tungsten that can be considered opaque. To make the input of the AAE model intuitive and generic, binary values are used to indicate the material selected for each layer, with ``0'' and ``1'' representing the Ge and SiO$_2$ layers, respectively. Each unit layer of the multilayer structure with a thickness of 0.11 $\mu m$. Following this way, 36-dimensional vectors represent different possible multilayer structures. According to the combinatorial theory, the total number of candidate structures is $2^{36}$, which composes a huge candidate structure space. The goal of the optimization is to determine the selection of the material sequence that enables the target thermal emission properties with a working band ranging from 4 to 7 $\mu m$.
The AAE generative network consists of an autoencoder and a discriminator, as shown in Fig. \ref{fig:1}(a), and its ability of dimensionality reduction and data interpretation \cite{creswell2018generative} allows a much broader variety of structure designs. The encoder ($E$) first compresses the input 36-dimensional vector parameter ($x$) into low-dimensional latent space. The decoder ($G$) reconstructs the real design parameters ($\tilde{x}$) based on the coordinates in the latent space $\tilde{z} = E(x)$. The encoder network in output layer has fewer nodes than in the input layer, so a compressed representation of input structural parameters can be learned. The autoencoder learning process uses binary cross entropy as a loss function. In addition, the discriminator ($D$) assures the latent space distributions $q(\tilde{z})$ to approach a predefined Gaussian distribution $p(z)$. Finally, the output of the encoder obtains a continuous space represented by compressed features. The loss functions to be minimized during the training of discriminator $L_{dis}$ (based on the min max game between $E$ and $D$ networks) and autoencoder $L_{aut}$ are calculated by (see Supplement 1 for details on the implementation of the AAE network),
\begin{equation}
L_{dis}=\min _{E} \max _{D}[\log (D(E(x)))+\log (1-D(\tilde{z}))]
\label{eq:1}
\end{equation}
\begin{equation}
L_{aut}=-\min _{E, G}[\log p(x \mid G(E(x)))].
\label{eq:2}
\end{equation}
After obtaining the compressed continuous and compact latent space, the brute-force approach and traditional optimization algorithms are still inefficient to search the optimal solution since it is a non-convex optimization problem. To realize high efficient global search and obtain maximum FOM, the AAE approach was further combined with BO. The BO is a statistical and probability-based algorithm to find the global optimal of black-box objective function which is expensive to evaluate. During the optimization process, BO uses the Gaussian process to construct the posterior distribution as a surrogate model to describe the objective function. The potential optimal solution is then suggested according to the acquisition function, and the decoder takes the initial randomly selected or BO recommended latent vectors as inputs and generates new multilayer structures with potential higher FOM, as shown in Fig. \ref{fig:1}(b). The dielectric functions of Ge, SiO$_2$ and W were obtained from Palik \cite{palik1998handbook}. The incident electric field are under p-polarized waves, and continuous boundary conditions were imposed to obtain the thermal emission spectra. To guarantee the target narrowband thermal radiation property, the thermal emission of the emitter should have a sharp emissivity at the target wavelength and zero emissivity elsewhere, here we define the FOM for each candidate structure to compare the differences between the calculated spectrums and the ideal emitter spectrums, 
\begin{equation}
FOM=\frac{\int_{\lambda_{1}}^{\lambda_{2}} \varepsilon E_{b} d \lambda}{\int_{\lambda_{1}}^{\lambda_{2}} E_{b} d \lambda}-\frac{\int_{\lambda_{1}}^{\lambda_{min}} \varepsilon E_{b} d \lambda}{\int_{\lambda_{1}}^{\lambda_{min}} E_{b} d \lambda}-\frac{\int_{\lambda_{2}}^{\lambda_{max}} \varepsilon E_{b} d \lambda}{\int_{\lambda_{2}}^{\lambda_{max}} E_{b} d \lambda}-P
\label{eq:3}
\end{equation}
where $\varepsilon$ denotes the spectral emissivity of each structure, which is the absorption $\alpha$ in the thermal equilibrium state according to Kirchhoff's law. $E_b$ is the spectral radiance density at a given wavelength $\lambda$ and temperature $T$. $\lambda_1$ and $\lambda_2$ are the minimum and maximum target wavelength, and $(\lambda_1-\lambda_2)/2$ was set to 4 $nm$. $\lambda_{min}$ and $\lambda_{max}$ are 4 and 7 $\mu m$, representing the lower and upper boundaries of the working band, respectively. The penalty term, $P = 0.1 \times$ the number of peaks, is added to make the emission spectrum smooth. The multilayer structure design problem finally becomes to minimize the difference between the designed emissivity spectrum and the target emitter property.

For narrowband thermal emitters with target wavelength at 4.5, 5.5, and 6.5 $\mu m$, 118, 157, and 136 pairs of multilayer structures and their corresponding emission spectra are prepared as initial training set for AAE network. The FOM distribution of the training set ranges from 0.28 to 0.81. To save time in preparing the training set, the jitter and scaling data augmentation method \cite{iwana2021empirical} is adopted to increase the training set to more than 20000 pairs (see Supplement 1 for data augmentation details). Once the AAE network is trained, it can generate numerous high quality designs within a few seconds. By combining the decoder with the optimization algorithm, a low-dimensional global optimization can be quickly achieved instead of a high-dimensional optimization. During the optimization, we randomly selected 300 initial candidate structures and calculated their emission spectrum by RCWA to evaluate each FOM. The 300 pairs of data were then used for the Bayesian prediction model to learn the posterior distribution of the black-box objective function. The Bayesian prediction model will then recommend 2D latent vectors that are more likely to achieve the target narrowband properties. This is based on the acquisition function which was set in the range of (-10, 10) to contain the majority of the latent space with standard normal distribution. By feeding the latent vector into the decoder of the trained autoencoder and combining it with the set threshold, a 36 dimensions binary matrix can be generated, which corresponds to the material selection order of the 36-layer structure. After accurately calculating the FOM of suggested structures by RCWA, the posterior distribution of the objective function will be updated. The model will then make the next round of recommendations based on the new acquisition function. Finally, the global optimal structural design can be quickly found by repeating this process. To avoid being trapped in local optimum during the search process, several random BO experiments were carried out and each time the total number of calculated structures was set as 2000. The Bayesian optimization package developed by Nogueira et al. \cite{nogueira2014bayesian} was used. 
\begin{figure}[ht!]
\centering
\includegraphics[width=\linewidth]{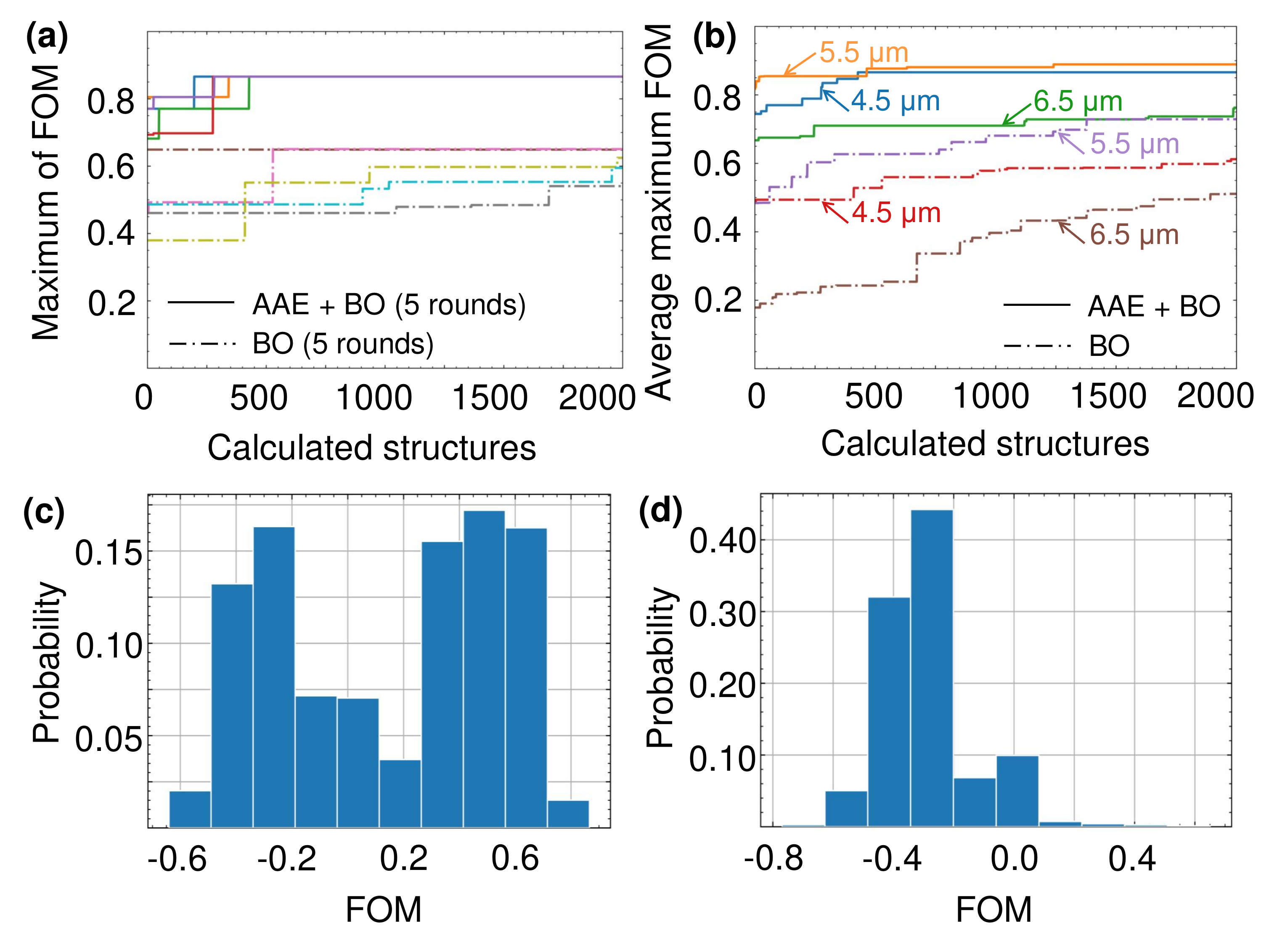}
\caption{(a) Optimization history of AAE+BO (solid line) and BO (dash dot line). Each line type contains 5 rounds stochastic optimization. (b) Average maximum FOM history of AAE+BO (solid line) and BO (dash dot line) for three target wavelengths. FOM distributions for 5 rounds of AAE+BO (c) and BO (d).}
\label{fig:2}
\end{figure}
Figure \ref{fig:2}(a) shows the optimization history of the maximum FOM with respect to the number of calculated structures for thermal emitter with target wavelength of 4.5 $\mu m$. To demonstrate the efficiency of AAE assisted BO design framework, we also used BO without the assistant of AAE network. 5 rounds stochastic AAE+BO and 5 rounds stochastic BO with different random seeds were conducted under the same parameter settings for fair comparison. The result shows that AAE+BO framework starts with a higher FOM, and all rounds of optimizations eventually convergence to 0.87. On the contrary, the pure BO starts with a lower FOM and the maximum FOM obtained at the end of the optimization is much smaller than the maximum FOM found by AAE+BO framework within 500 iterations. This comparison shows that the efficiency of the hybrid optimization framework is significantly higher than using pure BO. This solves one bottleneck problem of BO that it can hardly deal with high dimensional optimization problems due to the huge memory consumption. What's more, the total number of calculated candidate structures is significantly reduced to obtain the optimal result. To explore more about the optimization process, the histogram of the explored FOM distribution for 5 rounds of AAE+BO within 2000 iterations is shown in Fig. \ref{fig:2}(c). Around half of the FOMs are higher than the lowest FOM in the training data set, indicating again that the AAE+BO approach prefers to find structures with higher FOM during the optimization process. Fig. \ref{fig:2}(d) depicts the statistics of the FOM obtained from the 5 rounds of pure BO. Almost all discovered FOMs during the optimization process were below 0.2, illustrating the low efficiency of direct searching by BO for high dimensions problems.
\begin{figure}[ht!]
\centering
\includegraphics[width=\linewidth]{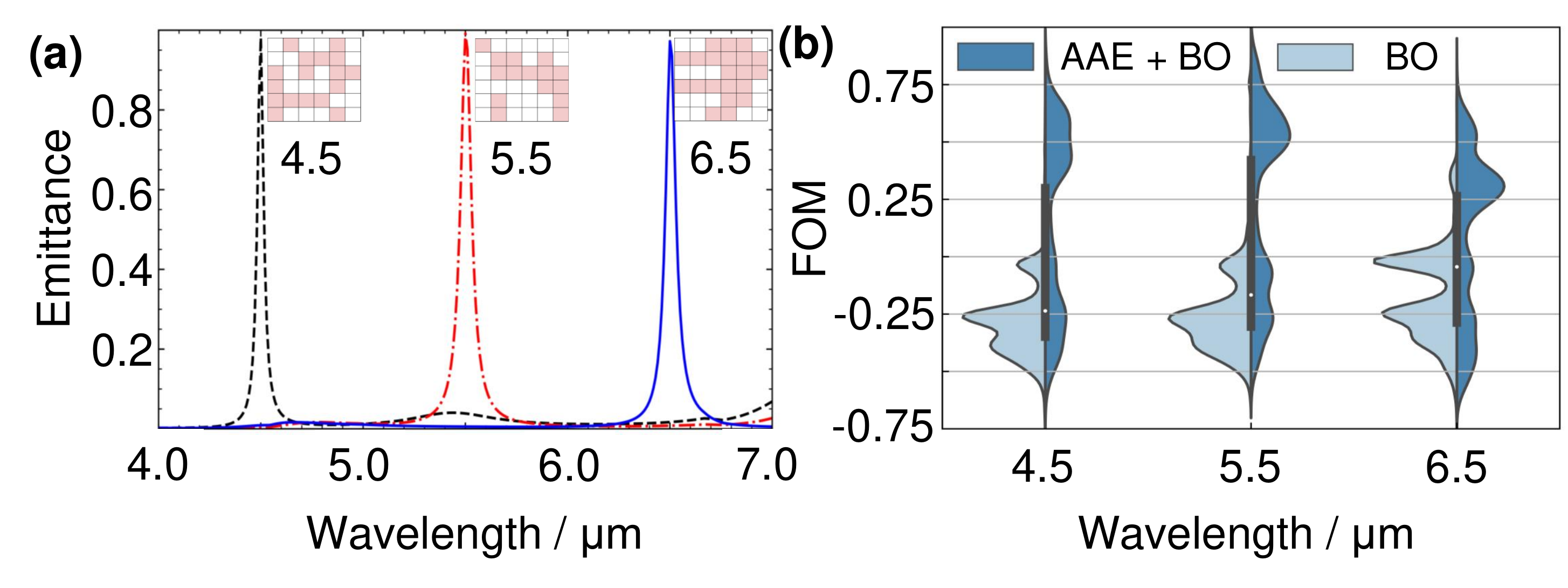}
\caption{(a) Emittance of the best designs searched by AAE+BO framework. The inset depicts the structure design corresponding to each emissivity. (b) Violin plot of the optimization process for the three target wavelengths. Light blue and dark blue represent the BO and AAE+BO framework, respectively.}
\label{fig:3}
\end{figure}
To exclude randomness, the convergence of average maximum FOM versus the calculated number of structures for cases with different target wavelengths was shown in Fig. \ref{fig:2}(b). It shows that the average maximum FOM designed by AAE+BO framework is at least 0.1 higher than that of the BO framework for the same target optimization. Nevertheless, the final average maximum FOM below 0.6 for AAE+BO framework at the 6.5 $\mu m$ case could be caused by the poor FOM distribution in the training set, which means that the trained AAE network has difficulty to learn the design principles for higher FOM structures. Fig. \ref{fig:3}(a) shows the emittance of the best multilayer structures designed by AAE+BO framework for three different target wavelengths, where the dashed, dash dot and solid curves represent the optimal designs with FOM of 0.87, 0.89 and 0.84, respectively. The inset of Fig. \ref{fig:3}(a) shows the corresponding material selection sequence reshaped to $6 \times 6$ pixels. According to the optimization result, the high emission at target wavelength and the rapid decay of emission at the non-target wavelength are realized. Although there is still room for further improving the emission spectra, the structures found here already satisfy the design target with less than 10,000 calculations, and the optimization efficiency is considerable for a search problem with candidate space of 68.7 billion. In addition, less than 2000 iterations per round enables to prevent the exponential increase of time for BO to store the computed structural information. Fig. \ref{fig:3}(b) shows the comparison of the FOM kernel density estimation with two optimization frameworks for different optimization cases. The thermal emitter designed by the BO tend to search structures with FOM less than 0. In contrast, the AAE+BO framework enables the search of new structures with FOM concentrated within $(0.2, 0.9)$. This analysis shows that the hybrid optimization framework can improve the search efficiency for on-demand design with a huge candidate space, and it is particularly important to train the AAE network to learn the connection between thermal emitter design and optical response.
\begin{figure}[ht!]
\centering
\includegraphics[width=\linewidth]{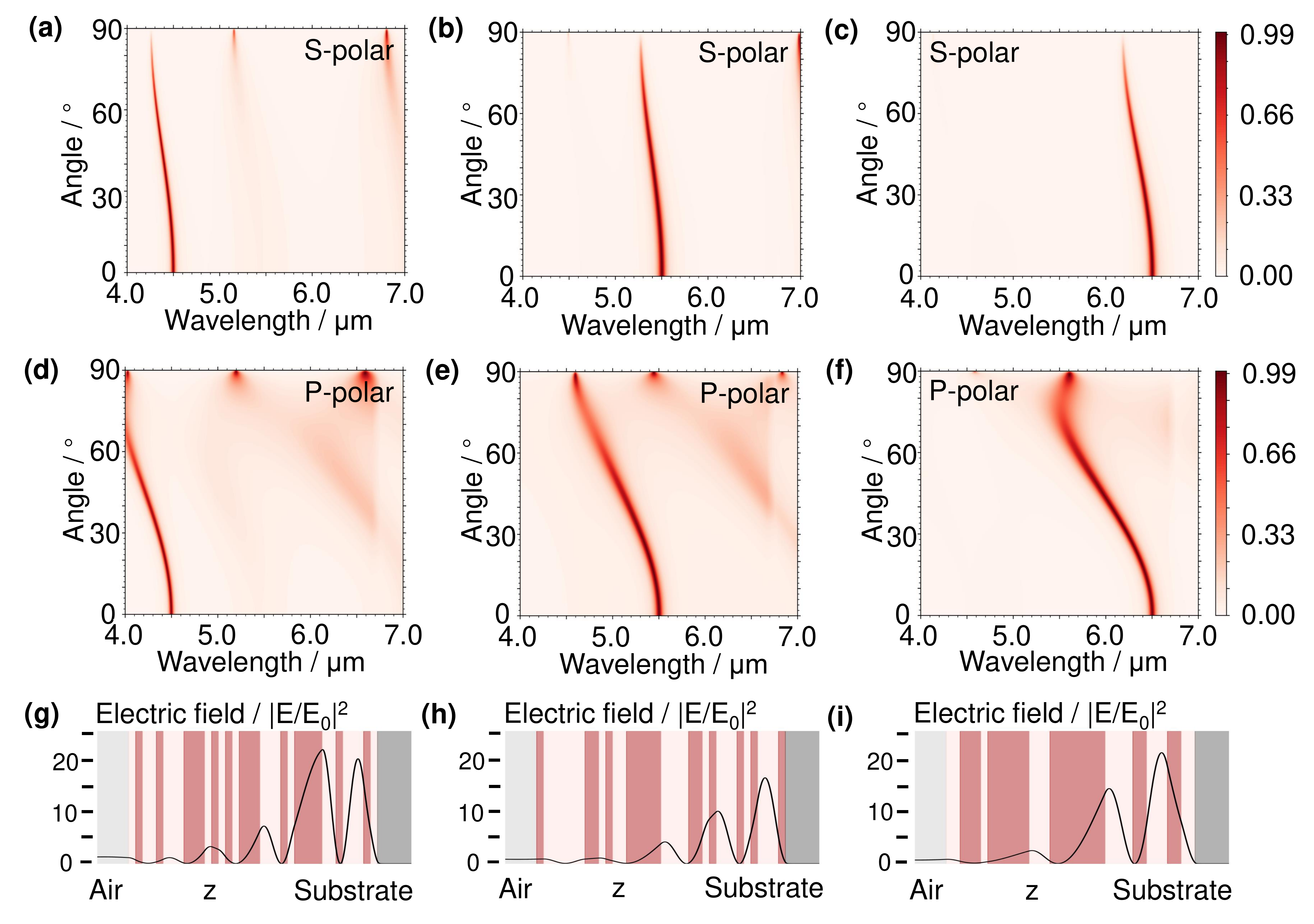}
\caption{(a)-(c) and (d)-(f) Angle-dependence emission spectra of optimal structures for s-polarized and p-polarized, respectively. (g)-(i) The normalized electric field amplitudes excited by a normally incident plane wave from air at the emission peak wavelength of 4.5, 5.5 and 6.5 $\mu m$, respectively.}
\label{fig:4}
\end{figure}
To understand the physical mechanisms behind the designed optimal structures, we further discuss the spectral-directional emission properties of the three optimal thermal emitters with different target wavelengths. As shown in Fig. \ref{fig:4}(a)-(f), the emission peak position is curved for large incident angles, especially in the case of p-polarized incident wave, which is caused by the localized mode. To further illustrate the mechanism of the enhanced emission at target wavelength, the electric field distribution inside the structures are shown in Fig. \ref{fig:4}(g)-(i). The x-axis refers to the thickness of multilayer structures, the y-axis to the electric field intensity, and the optimal structure is excited by a normally incident plane wave. Fig. \ref{fig:4}(g) and (i) indicates that the electric fields are strongly enhanced inside the Ge layer. The last four layers of the 4.5 $\mu m$ structure and the last two layers of the 6.5 $\mu m$ structure can be seen as cavities. Besides, both structures exhibit exponentially oscillating decay, which means the coupling of Fabry-Perot resonance and Tamm mode, and structures will have a smaller line width. As for the target wavelength of 5.5 $\mu m$ shown in Fig. \ref{fig:4}(h), the designed aperiodic structure leads to an exponential decay, which can be characterized as the Tamm mode serves to a sharp peak emissivity.

In conclusion, we have developed a hybrid optimization framework that combines the AAE network with BO to design wavelength-selective narrowband thermal emitters. With several hundreds of training data sets that roughly satisfy the target properties, the hybrid model enables the novel optimal structure design, which greatly saves the global search cost. This benefit comes from the difference of original high-dimensional degrees of freedom and low-dimensional latent space. Comparing the optimization by pure BO, it further demonstrates the effectiveness, feasibility, and accuracy of the proposed design framework, which greatly extends the application of BO to high-dimensional design cases. The analysis of the physical mechanisms make it possible to gain new expert knowledge regarding the electrodynamics mode components that lead to the optimal metamaterials. The proposed hybrid optimization framework have potential applications in thermal radiation and beyond.




\begin{acknowledgments}
This work was supported by the Shanghai Pujiang Program (No. 20PJ1407500), National Natural Science Foundation of China (No. 52006134), Shanghai Key Fundamental Research Grant (No. 21JC1403300), JSPS KAKENHI (No. 19K14902), Materials Genetic Engineering Project for Rare and Precious Metals by Yunnan Province (Grant No. 202002AB080001-1), and Opening Project of State Key Laboratory of Green Building Materials (Grant No. 2020GBM02). The computations in this work were run on the $\pi$ 2.0 cluster sup-ported by the Center for High Performance Computing at Shanghai Jiao Tong University. The computation was also partially supported by Open Research Fund of CNMGE Platform \& NSCC-TJ. 

\end{acknowledgments}

\end{document}